\documentclass[10pt,twocolumn]{article} 
\usepackage{simpleConference}
\usepackage{times}
\usepackage{graphicx}
\usepackage{amssymb}
\usepackage{url,hyperref}
\usepackage{cite}
\usepackage{amsmath,amssymb,amsfonts}
\usepackage{algorithmic}
\usepackage{graphicx}
\usepackage{textcomp}
\usepackage{subfigure} 
\usepackage{stfloats}
\usepackage{amsmath}
\usepackage{booktabs}
\usepackage{multirow}
\usepackage{url}
\usepackage[font=small,labelfont=bf]{caption}
\begin{document}

\title{Block Shuffle: A Method for High-resolution Fast Style Transfer with Limited Memory}

\author{Weifeng Ma, Zhe Chen, and Caoting Ji \\
\\
School of Information and Electronic Engineering \\
Zhejiang University of Science and Technology \\
Hangzhou, China \\
\today
\\
\\
mawf@zust.edu.cn  \\
}

\maketitle
\thispagestyle{empty}

\begin{abstract}
Fast Style Transfer is a series of Neural Style Transfer algorithms that use feed-forward neural networks to render input images. Because of the high dimension of the output layer, these networks require much memory for computation. Therefore, for high-resolution images, most mobile devices and personal computers cannot stylize them, which greatly limits the application scenarios of Fast Style Transfer. At present, the two existing solutions are purchasing more memory and using the feathering-based method, but the former requires additional cost, and the latter has poor image quality. To solve this problem, we propose a novel image synthesis method named \emph{block shuffle}, which converts a single task with high memory consumption to multiple subtasks with low memory consumption. This method can act as a plug-in for Fast Style Transfer without any modification to the network architecture. We use the most popular Fast Style Transfer repository on GitHub as the baseline. Experiments show that the quality of high-resolution images generated by our method is better than that of the feathering-based method. Although our method is an order of magnitude slower than the baseline, it can stylize high-resolution images with limited memory, which is impossible with the baseline. The code and models will be made available on \url{https://github.com/czczup/block-shuffle}.
\end{abstract}

\section{Introduction}
\label{sec:introduction}
Fast Style Transfer\cite{johnson2016perceptual,ulyanov2016texture} uses feed-forward neural networks to learn artistic styles from paintings and uses the learned style information to render input images. This technology improves the speed of Gatys \emph{et al.}'s algorithm\cite{gatys2016image} and promotes the industrialization of Neural Style Transfer. For example, Prisma\cite{prisma} is a famous mobile application based on Fast Style Transfer. It has set off the trend of using photos for artistic creation, and more and more people are enthusiastic about using this application to render their photos and share them on social networks. For such a simple application scenario, it does not require high-resolution images. However, in recent years, people try to apply Fast Style Transfer to some new scenarios, such as customizing decorative paintings, making video special effects, and synthesizing art posters. Unlike sharing photos on social networks, these new application scenarios need to stylize high-resolution images.

However, due to memory limitations, most ordinary devices are unable to stylize high-resolution images directly. Specifically, Fast Style Transfer includes a feed-forward neural network for image transformation and a pre-trained network for loss calculation. The image transformation network is a fully convolutional neural network, which can process images of arbitrary size. But in practice, the maximum resolution of the input image is determined by the memory of the device, and oversized images will cause out-of-memory (OOM) errors.

There are two existing solutions to this problem. One is to buy more memory to meet computing needs, but this approach increases the cost and does not completely solve the problem. Another one is to divide the input image into many overlapping sub-images, stylize them respectively, and then use the feathering effect \cite{ghosh2016survey, li2008automatic} to stitch them (hereinafter referred to as feathering-based method). This method does not need to upgrade the hardware, but its output image has obvious seamlines. For example, Painnt\cite{painnt} is a mobile application that uses the feathering-based method to stylize high-resolution images locally.

To solve the problems in the above two methods, we delve into the characteristics of Fast Style Transfer models and propose a novel method named \emph{block shuffle}. Its main contributions are as follows:
\begin{enumerate}
	\item This method converts a single task with high memory consumption to multiple subtasks with low memory consumption. It enables more ordinary devices to support high-resolution style transfer, extending the scope of application of Fast Style Transfer.
	\item Compared with the feathering-based method, our method eliminates the seamlines and small noise textures, which significantly improves the quality of generated images.
	\item This method is non-invasive, which only adds pre-processing and post-processing steps before and after the image transformation network, and does not need to retrain the model.
\end{enumerate}

\begin{figure*}[ht]
	\centering
	\includegraphics[width=17.4cm]{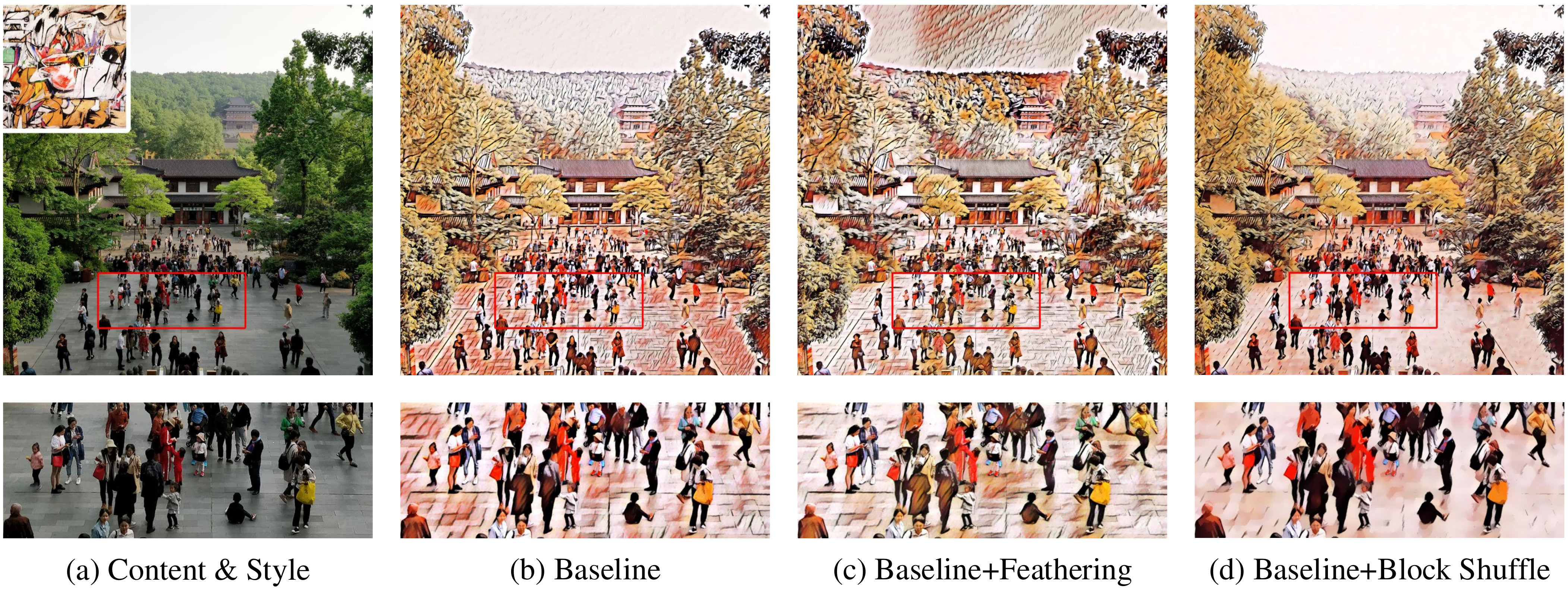}
	\caption{Comparison of baseline, baseline+feathering-based method, and baseline+block shuffle (ours). The resolution of the above four images is all $3000\times 3000$. The baseline is the most popular Fast Style Transfer repository on GitHub\cite{engstrom2016faststyletransfer}.}
	\label{fig1} 
\end{figure*}

\section{Related Work}
In 2016, based on the previous work of texture synthesis \cite{gatys2015texture}, Gatys \emph{et al.} first proposed a Neural Style Transfer algorithm \cite{gatys2016image}. By reconstructing representations from the feature maps in the VGG-19 network \cite{simonyan2014very}, they found it has a strong feature extraction capability. Its lower layers can capture the content information of the input image, and its upper layers can capture the style information of the input image. Therefore, they designed content loss and style loss based on the VGG-19 network and achieved high stylization quality. However, their method is based on online image optimization, and each style transfer requires several hundred iterations, which takes a long time.

In order to speed up the process of Neural Style Transfer, Johnson \emph{et al.}\cite{johnson2016perceptual} and Ulyanov \emph{et al.} \cite{ulyanov2016texture} respectively proposed a method of training a feed-forward neural network, which can stylize the input image only through a forward pass. Images generated by their methods are similar to that of Gatys \emph{et al.}'s method \cite{gatys2016image}, but the speed is three orders of magnitude faster, so these methods are collectively called Fast Style Transfer. For example, using an Nvidia Quadro M6000 GPU to stylize a 512$\times$512 image, the method of Gatys \emph{et al.} takes 51.19 seconds, while the methods of Johnson \emph{et al.} and Ulyanov \emph{et al.} only take 0.045 seconds and 0.047 seconds\cite{jing2019neural}. 

In addition to speed, researchers have made many improvements in the quality of style transfer. For example, Ulyanov \emph{et al.}  proposed the instance normalization \cite{ulyanov2016instance}, which applies normalization to every single image rather than a batch of images. Using instance normalization instead of batch normalization \cite{ioffe2015batch} can not only promote convergence but also significantly improve the quality of generated images. Besides, Gatys \emph{et al.} reviewed their previous style transfer algorithm \cite{gatys2016image} and found that the stroke size is related to the receptive field of the VGG-19 network. For a high-resolution image, the receptive field is much smaller than the image, so this algorithm cannot produce large stylized structures. Therefore, they proposed a coarse-to-fine method that could generate high-resolution images with large brush strokes \cite{gatys2017controlling}.

At present, the research of high-resolution Fast Style Transfer mainly focuses on the brush strokes, and there is no research to solve the uncomputability problem due to limited memory in practical application. For example, Wang \emph{et al.} \cite{wang2017multimodal}, Zhang \emph{et al.} \cite{zhang2018multi} and Jing \emph{et al.} \cite{jing2018stroke} all studied the brush strokes of Fast Style Transfer, aiming at producing excellent high-resolution images. These methods enlarge the stroke size, but they cannot process oversized images due to the limitation of device memory. Therefore, we propose a novel method named \emph{block shuffle}, which not only solves this problem effectively but also can produce higher quality images than the feathering-based method.

\begin{figure*}[ht]
	\centering
	\includegraphics[width=17.4cm]{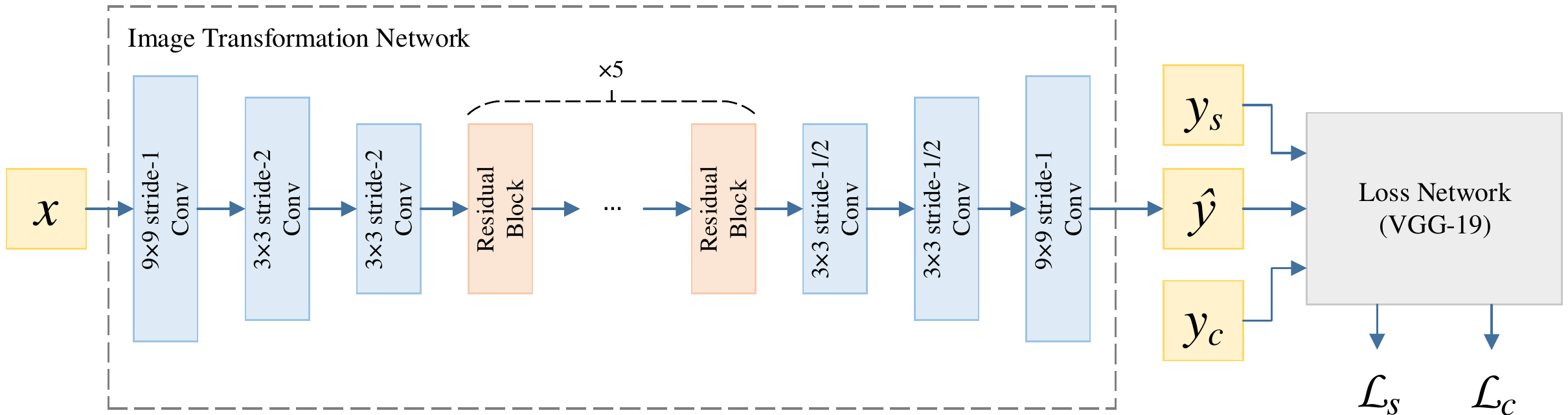}
	\caption{The model architecture of the baseline. $x$ is the input image, $\hat{y}$ is the output image, $y_s$ is the style image, and $y_c$ is the content image ($x = y_c$).}
	\label{fig2} 
\end{figure*}

\section{Pre-analysis}
We use the most popular Fast Style Transfer repository \cite{engstrom2016faststyletransfer} on GitHub as the baseline. In this section, we will briefly introduce the model architecture and loss function of the baseline and analyze the reasons for the poor performance of the feathering-based method.
\subsection{Model Architecture}
Based on previous research, Engstrom implemented Fast Style Transfer and shared source code on GitHub \cite{engstrom2016faststyletransfer}, which attracts many developers and researchers. As shown in Fig. \ref{fig2}, in this repository, the loss network is a VGG-19 network pre-trained on the ImageNet dataset \cite{deng2009imagenet}, and the image transformation network is a 16-layer deep residual network. 

The architecture of the image transformation network is as follows: The kernel size of the first and last convolutional layers is 9$\times$9, and that of others is 3$\times$3. The second and third layers are stride-2 convolutions, which are used for downsampling. The last two and three layers are fractionally-strided convolutions with stride 1/2, which are used for upsampling (i.e., transposed convolutions with stride 2). The middle ten layers are composed of five residual blocks \cite{he2016identity}, and each residual block contains two convolutional layers. All non-residual convolutional layers are followed by instance normalization and ReLU activation function.

\subsection{Loss Function}
The loss function in the baseline combines the design of Gatys \emph{et al.} \cite{gatys2016image} and Johnson \emph{et al.} \cite{johnson2016perceptual}, which consists of three parts: style loss $\mathcal{L}_s$, content loss $\mathcal{L}_c$ and total variation loss $\mathcal{L}_{tv}$. The total loss is expressed as:
\begin{equation}
\mathcal{L}(\hat{y},y_c,y_s)=\lambda_s\mathcal{L}_s(\hat{y},y_s)+\lambda_c\mathcal{L}_c(\hat{y},y_c)+\lambda_{tv}\mathcal{L}_{tv}(\hat{y})
\end{equation}
where $\lambda_s$, $\lambda_c$, and $\lambda_{tv}$ are the tradeoff parameters for style loss, content loss, and total variation loss.
\subsubsection{Style Loss}
The style loss $\mathcal{L}_s$ is used to measure the style consistency between the output image $\hat{y}$ and the style image $y_s$. First, input $\hat{y}$ and $y_s$ to the VGG-19 network, then take the feature maps of layers $relu1\_1$, $relu2\_1$, $relu3\_1$, $relu4\_1$,  and $relu5\_1$ to compute the Gram matrix respectively, and finally calculate the Euclidean distance between the Gram matrix of these two images:
\begin{equation}
\mathcal{L}_s(\hat{y},y_s) = \sum_{l\in layers}\left \| \mathcal{G}(\mathcal{F}_l(\hat{y}))-\mathcal{G}(\mathcal{F}_l(y_s)) \right \|^2
\end{equation}
where $\mathcal{F}_l(\cdot)$ represents the feature maps of layer $l$ in the VGG-19 network, and $\mathcal{G}(\cdot)$ represents the Gram matrix. When computing the Gram matrix, reshape the feature maps $\mathcal{F}_l(\cdot)$ of shape $C_l\times H_l\times W_l$ into a matrix $\psi$ of shape $C_l\times H_l W_l$, then $\mathcal{G}(\mathcal{F}_l(\cdot))=\psi\psi^T/C_l W_l H_l$.
\subsubsection{Content Loss}
The content loss $\mathcal{L}_c$ is used to measure the content consistency between the output image $\hat{y}$ and the content image $y_c$. First, input $\hat{y}$ and $y_c$ into the VGG-19 network, and then take feature maps of layer $l=relu3\_3$ to compute the Euclidean distance:
\begin{equation}
\mathcal{L}_c(\hat{y},y_c) = \frac{1}{C_l W_l H_l}\left \| \mathcal{F}_l(\hat{y})-\mathcal{F}_l(y_c)) \right \|^2
\end{equation}

\subsubsection{Total Variation Loss}

Total variation loss $\mathcal{L}_{tv}$ can promote the model to produce a smooth image, which is defined as:
\begin{equation}
\mathcal{L}_{tv}(x) = \sum_{i,j} \left| x_{i+1,j}-x_{i,j} \right|+\left| x_{i,j+1}-x_{i,j} \right|
\end{equation}
where $x_{i,j}$ is a pixel on image $x$, and $i,j$ represent the position of this pixel.

\begin{figure*}[ht]
	\centering
	\includegraphics[width=17.4cm]{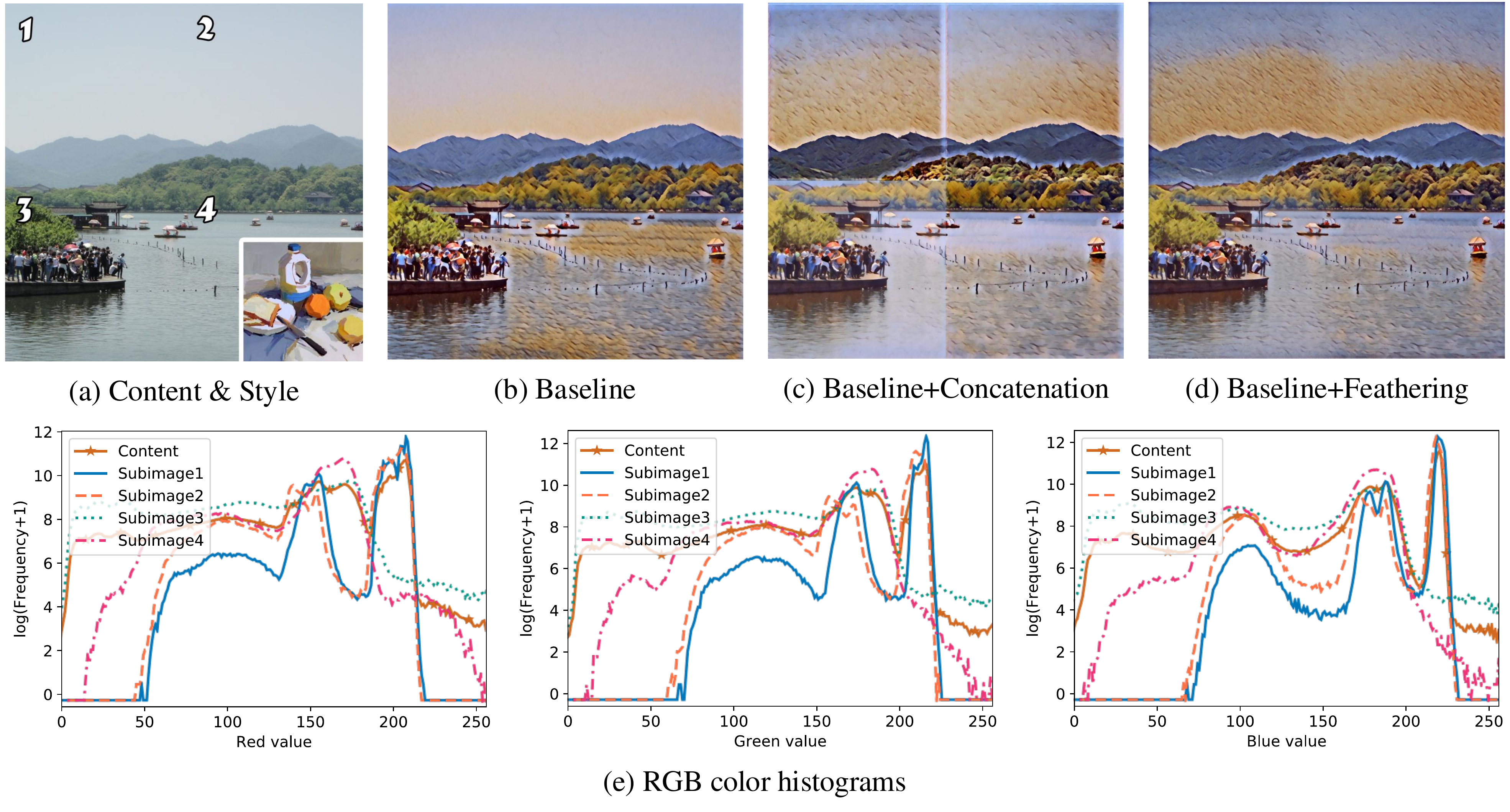}
	\caption{Problem analysis. The resolution of the above four images is all $2000\times 2000$. }
	\label{fig3} 
\end{figure*}

\subsection{Problem Analysis}
\label{sec:Problem analysis}

On devices with limited memory, the image transformation network cannot stylize high-resolution images directly. Therefore, we need to divide the input image into many sub-images for processing (Fig. \ref{fig3}(a)). For example, we can cut the input image $x$ into many non-overlapping sub-images, stylize them respectively, and concatenate them to generate a complete image (Fig. \ref{fig3}(c)). However, this method results in a significant difference between the two adjacent stylized sub-images, which destroys the visual integrity of the output image. To improve this method, an intuitive idea is to generate overlapping sub-images and use the feathering effect to stitch them (i.e., feathering-based method), but this way still generates visible seamlines in the output image (Fig. \ref{fig3}(d)). Currently, mobile application Painnt has taken such a flawed method to stylize high-resolution images.

Observing the architecture of the image transformation network, we found two points that lead to this phenomenon: the receptive field of the image transformation network and the instance normalization layer. In convolutional neural networks, the receptive field is a region of the input image that affects a particular value in the feature maps of the network. Specifically, for a Fast Style Transfer model, a pixel on the output image $\hat{y}$ depends on the pixel distribution in the corresponding receptive field on the input image $x$. Besides, results of instance normalization also rely on the pixel distribution of the input image $x$. Therefore, in summary, the output image $\hat{y}$ will be affected by the pixel distribution of the input image $x$.

Based on the above analysis, we drew the RGB color histograms of the input image $x$ and its sub-images (Fig. \ref{fig3}(e)), from which we can observe that the pixel distribution of sub-images is quite different from that of the input image $x$. Therefore, we proposed a conjecture: if the pixel distribution of the sub-images matches that of the input image $x$, then its stylized results will also be similar and can be easily stitched.

\begin{figure*}[ht]
	\centering
	\includegraphics[width=17.4cm]{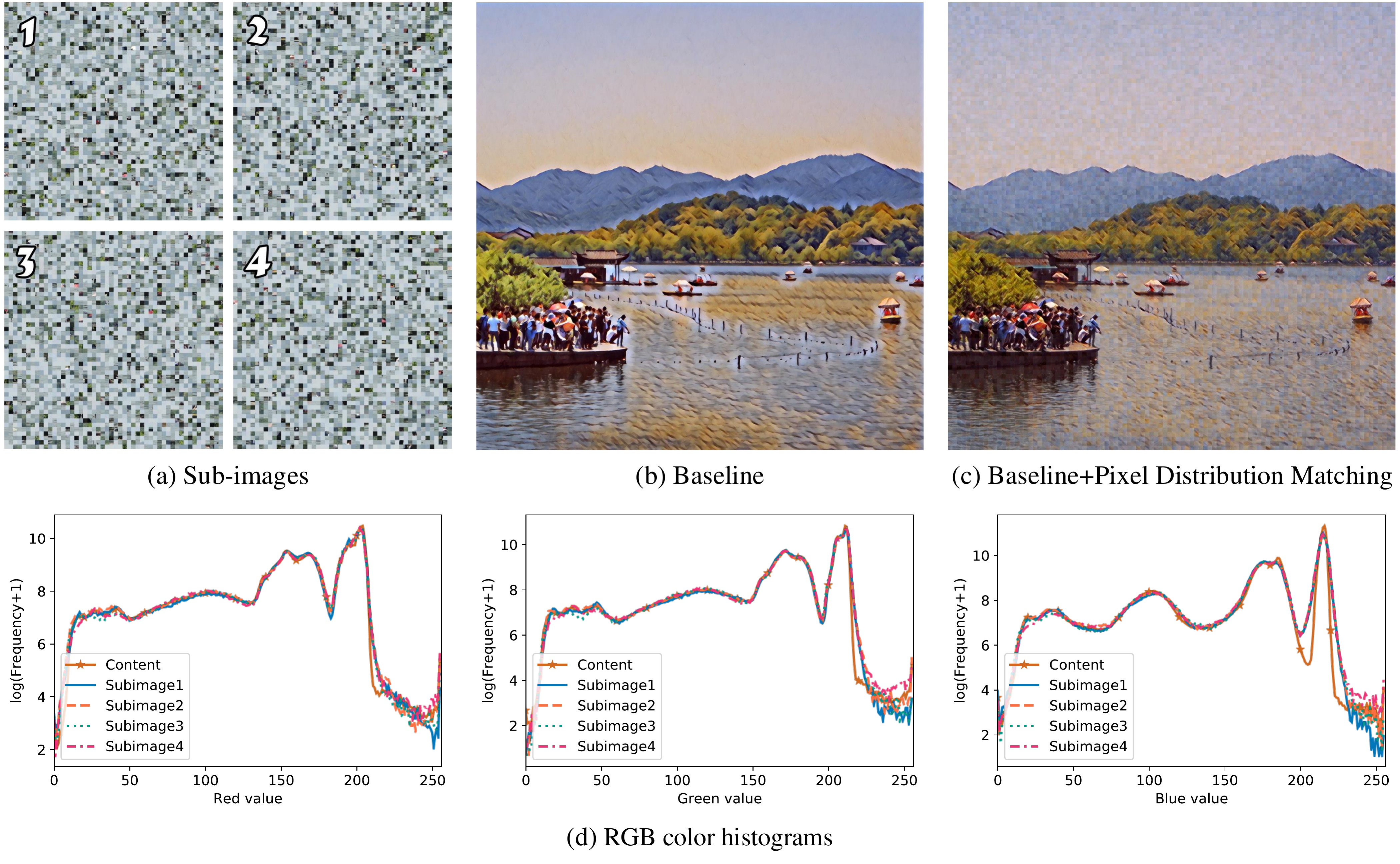}
	\caption{The pixel distribution matching method. The style image and content image are the same with Fig. \ref{fig3}. The resolution of these sub-images is all $1000\times 1000$ pixels. Image blocks in these sub-images are $20\times 20$ pixels.}
	\label{fig4} 
\end{figure*}

\section{Proposed Method}

In this section, we designed the \emph{pixel distribution matching} method, which proves the correctness of our conjecture. Based on that, we proposed the \emph{block shuffle} method.

\subsection{Pixel Distribution Matching}

In order to produce sub-images whose pixel distribution matches that of the input image $x$, we proposed the \emph{pixel distribution matching} method. First of all, we assume that the input image $x$ is a 3-channel image with width $W$ and height $H$, and then we process the input image $x$ by the following steps:

\begin{enumerate}
	\item Cut the input image $x$ into non-overlapping blocks of $w\times w$ pixels and number them in sequence (to simplify the discussion, we assume that both $W/w$ and $H/w$ are divisible).
	\item Shuffle the list of image blocks randomly and take out some image blocks every time to generate a sub-image.
\end{enumerate}

This method uses simple random sampling without replacement (SRSWOR) to select image blocks. In the population, each image block has an equal chance of getting selected, so the sub-images generated by this method (Fig. \ref{fig4}(a)) can better represent the input image $x$. As shown in Fig. \ref{fig4}(d), the pixel distribution of these sub-images is similar to that of the input image $x$. Besides, the above steps are like the patch shuffle regularization proposed by Kang \emph{et al.} \cite{kang2017patchshuffle}. It can be understood as a kind of regularization, which makes each sub-image contains not only local information but also global information of the input image $x$. 

Next, stylize all sub-images. Then, process the stylized sub-images by the following steps:
\begin{enumerate}
	\item Recut all stylized sub-images into image blocks of $w\times w$ pixels.
	\item Sort the list of image blocks according to their number.
	\item Concatenate all image blocks, then obtain an output image with width $W$ and height $H$.
\end{enumerate}

We observed the output image of this method (Fig. \ref{fig4}(c)) and found that the brightness and color of image blocks are slightly different. But overall, this output image is very similar to the result of the baseline (Fig. \ref{fig4}(b)). This phenomenon proves the correctness of the conjecture made in Section \ref{sec:Problem analysis} and provides theoretical support for further research.

\subsection{Block Shuffle}
\label{sec:blockshuffle}
\begin{figure*}[t]
	\centering
	\includegraphics[width=17.4cm]{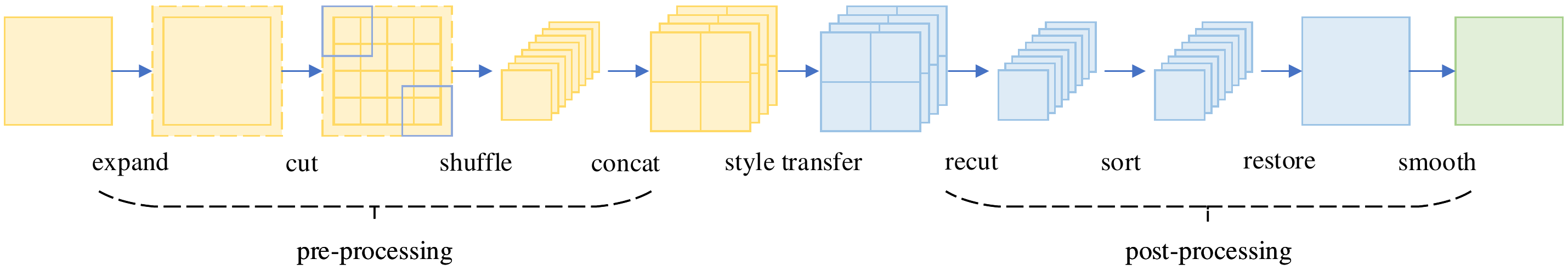}
	\caption{The processing flow of the block shuffle method.}
	\label{fig5} 
\end{figure*}

\begin{figure}[b]
	\centering
	\includegraphics[width=8cm]{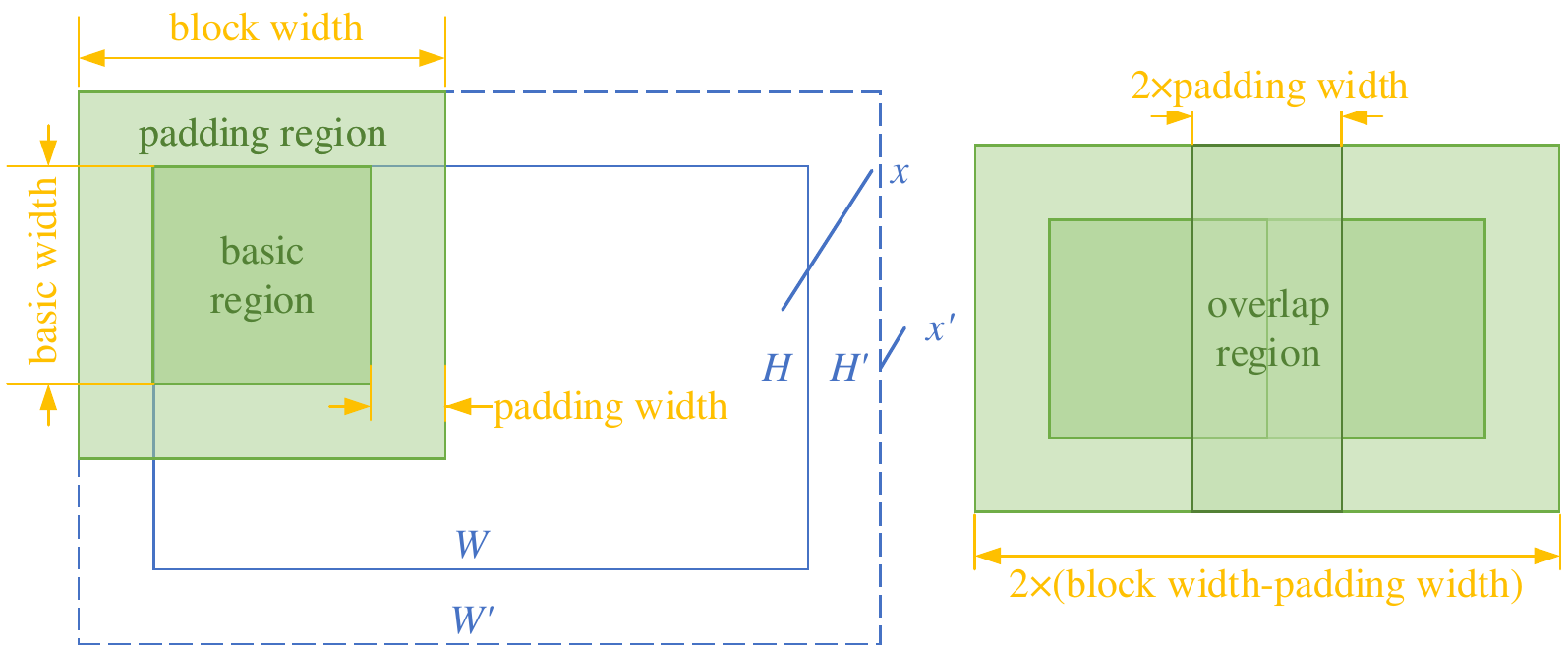}
	\caption{Definition of the image block.}
	\label{fig6}
\end{figure}

Based on the \emph{pixel distribution matching}, we propose the \emph{block shuffle} method, which improves the coherence of the output image. In this method, the four steps before the style transfer model are named pre-processing, and the four steps after that are named post-processing (as shown in Fig. \ref{fig5}). The specific process is as follows:

\textbf{Input parameters}. This method requires four input parameters: style transfer model $\mathcal{M}$, input image $x$, basic width $w_{basic}$, and padding width $w_{padding}$. As shown in Fig. \ref{fig6}, each image block is a square, consisting of a basic region and a padding region. For two adjacent image blocks, the overlapping part constitutes the "overlap region". The width of image blocks is expressed as:

\begin{equation}
w_{block}= w_{basic}+2w_{padding}
\end{equation}


\textbf{(1) Expand}. In order to ensure the integrity of image blocks, we use reflection padding to expand the input image $x$ from $W \times H$ to $W^{'} \times H^{'}$. The expanded image is represented as $x^{'}$, whose width $W^{'}$ and height $H^{'}$ are expressed as: 

\begin{equation}
\left\{
\begin{array}{lr}
W^{'}= \lceil W / w_{basic} \rceil \times w_{basic}+2w_{padding} \\
H^{'}= \lceil H / w_{basic} \rceil \times w_{basic}+2w_{padding} \\
\end{array} \right.
\end{equation}

\textbf{(2) Cut}. First, cut the image $x^{'}$ into overlapping square blocks with a width of $w_{block}$, and then number them in order. Specifically, we use a sliding window to crop the image, the size of the window is $w_{block}\times w_{block}$, and the stride of the window is $w_{basic}$. After that, the number of image blocks in the horizontal and vertical direction are respectively presented as $\lceil W / w_{basic} \rceil$ and $\lceil H / w_{basic} \rceil$, and the total number of blocks is:

\begin{equation}
N_{total}=\lceil W / w_{basic} \rceil \times \lceil H / w_{basic} \rceil
\end{equation}

\textbf{(3) Shuffle}. Shuffle the list of image blocks.

\textbf{(4) Concatenate}. Suppose our device can directly stylize an image of $w_{max} \times w_{max}$ pixels at most, so the size of sub-images must be less than or equal to this size. In the largest sub-image, the number of blocks is expressed as:
\begin{equation}
N_{block}=\lfloor w_{max} / w_{block} \rfloor ^2 
\end{equation}
Therefore, every time we take $N_{block}$ image blocks from the list in sequence and concatenate them into a square sub-image of $ (\sqrt{N_{block}} \times w_{block})\times (\sqrt{N_{block}} \times w_{block})$ pixels. The total number of sub-images is:
\begin{equation}
N_{subimg}=\lceil N_{total} / N_{block} \rceil 
\end{equation}

\textbf{(5) Style transfer}. Use the Fast Style Transfer model $\mathcal{M}$ to stylize all sub-images.

\textbf{(6) Recut}. First, recut the stylized sub-images into square image blocks with a width of $w_{block}$. Then, in order to reduce the boundary effect (i.e., the border area of stylized image blocks is contaminated by surrounding image blocks), remove the border area of 8 pixels wide around the image blocks, so the final width of image blocks is $w_{block}-16$.

\textbf{(7) Sort}. Sort the list of image blocks according to their number.

\textbf{(8) Restore}. First, use the feathering effect\cite{li2008automatic} to stitch all the image blocks. Then, remove the padding area added in step one and restore the image to its original size. Concretely, this feathering effect blends the left and right images by calculating the weighted average values in the overlap region:
\begin{equation}
p= \frac{d_l}{d_l+d_r}p_l+\frac{d_r}{d_l+d_r}p_r 
\end{equation}
where $p_l$ and $p_r$ are the pixels in the overlap region of the left image and the right image. $d_l$ and $d_r$ are the distance between the overlapping pixel and the border of the left and right images.

\textbf{(9) Smooth}. Finally, in order to eliminate the seamlines and small noise textures, we apply bilateral filters to the generated image, which smooth the image while preserving edges. To reduce the time spent, we use four small bilateral filters (sigmaColor=10, sigmaSpace=10) instead of a large bilateral filter (sigmaColor=40, sigmaSpace=40).

\begin{figure*}[ht]
	\centering
	\includegraphics[width=17.4cm]{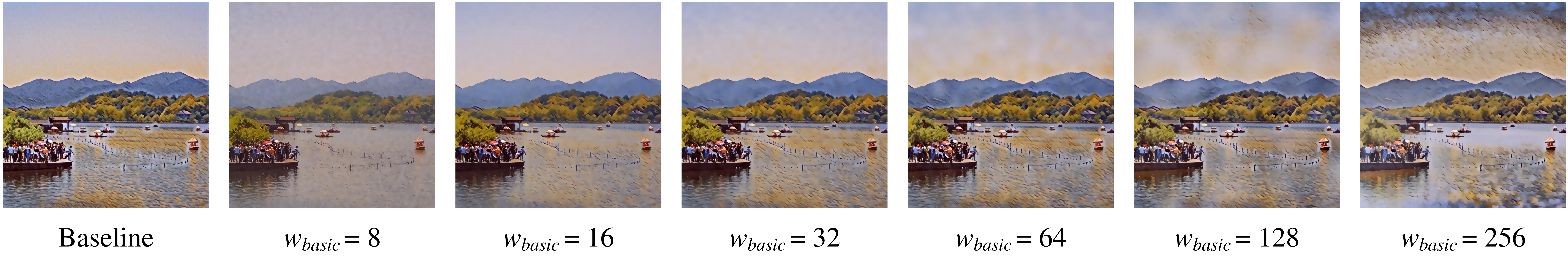}
	\caption{Comparison of different $w_{basic}$. The resolution of the above images is all $200\times 2000$.}
	\label{fig7} 
\end{figure*}

\begin{figure*}[!h]
	\centering
	\includegraphics[width=17.4cm]{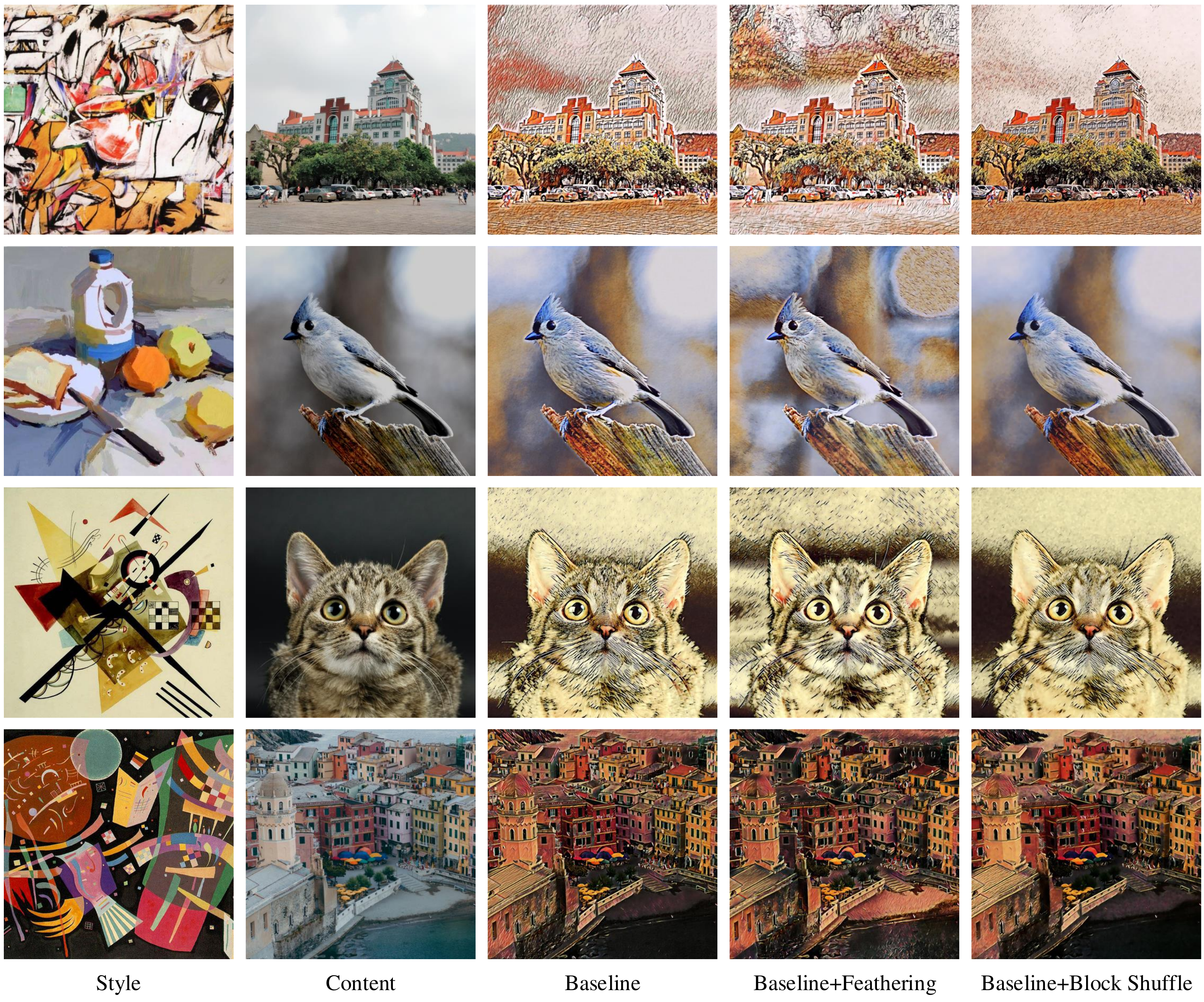}
	\caption{Comparison of baseline, baseline+feathering-based method, and baseline+block shuffle (ours). The resolution of the above images is all $3000\times 3000$.}
	\label{fig8} 
\end{figure*}

\section{Experiments and Result Analysis}

\subsection{Implementation Details}
In this paper, we adopt the most popular Fast Style Transfer repository \cite{engstrom2016faststyletransfer} on GitHub as the baseline. At training time, we used MS-COCO dataset\cite{lin2014microsoft} to train the network, and all the images are cropped and resized to $512\times512$ pixels. In addition, the Adam optimizer \cite{kingma2014adam} was used during training, with a learning rate of $1\times10^{-3}$. The batch size is 4, and the number of iterations is 40,000. The tradeoff parameters $\lambda_s$, $\lambda_c$, and $\lambda_{tv}$ are set to 100, 7.5, and 200, respectively. At test time, we use the baseline, the feathering-based method, and our block shuffle method to stylize high-resolution images. In our method, the maximum resolution $w_{max}\times w_{max}$ is set to $1000\times 1000$.

\subsection{Hyper-parameters Selection}
There are two crucial hyper-parameters in our block shuffle method, the basic width $w_{basic}$ and the padding width $w_{padding}$, which determine the structure of image blocks. According to the description in Section \ref{sec:blockshuffle}, the padding region will increase the calculation amount of style transfer. To estimate the computational complexity of our method compared to the baseline, we defined a parameter $\alpha$ as:

\begin{figure*}[ht]
	\centering
	\includegraphics[width=17.4cm]{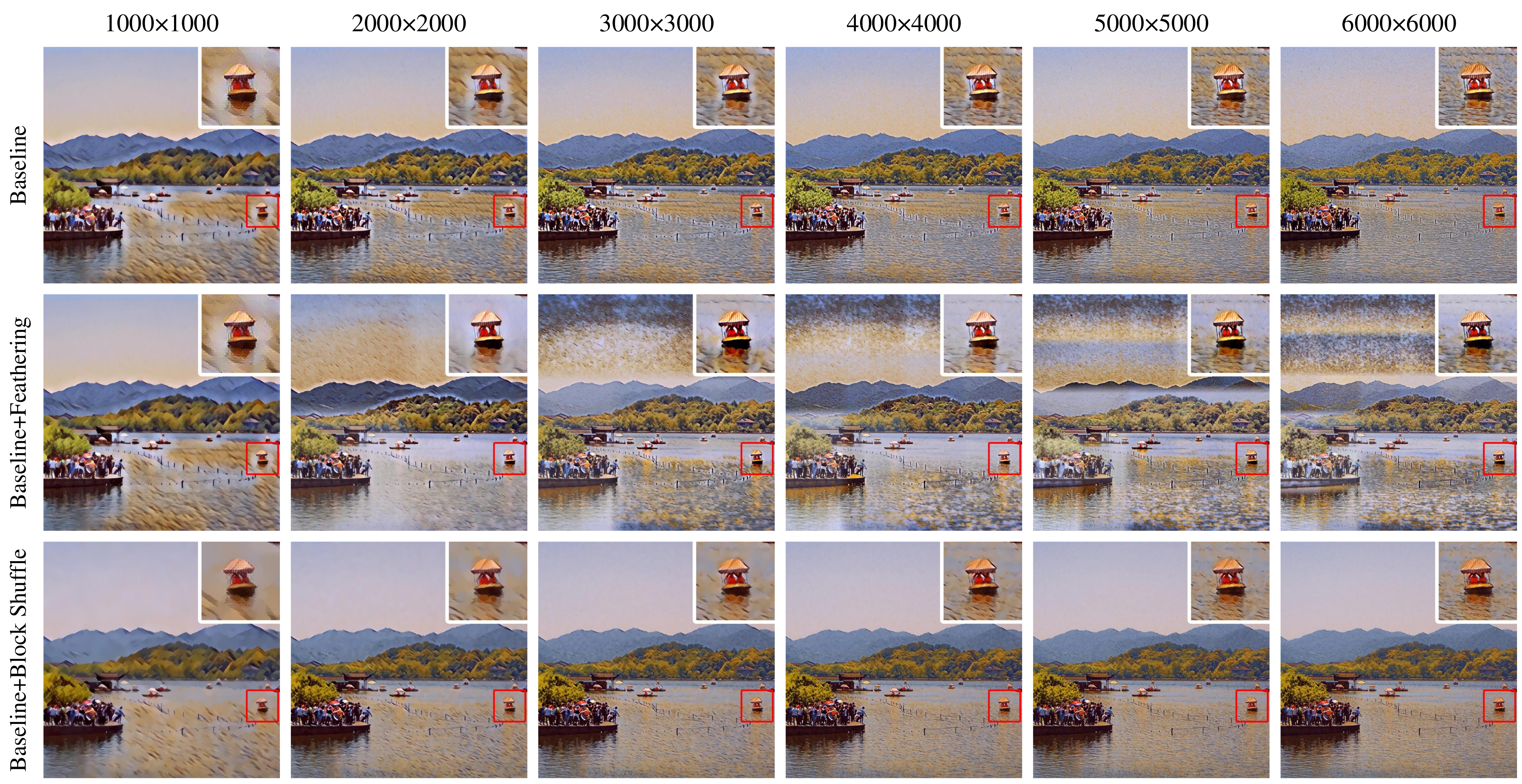}
	\caption{Comparison of baseline, baseline+feathering-based method, and baseline+block shuffle (ours) at different resolutions. The style image and content image are the same with Fig. \ref{fig3}.}
	\label{fig9} 
\end{figure*}

\begin{equation}
\begin{aligned}
\alpha=\left( \frac{w_{block}}{w_{basic}}\right)^2 &= \left( \frac{w_{basic}+2w_{padding}}{w_{basic}}\right)^2\\
&= \left(1+2\times \frac{w_{padding}}{w_{basic}}\right)^2
\end{aligned}
\end{equation}
where the ratio of $w_{padding}$ to $w_{basic}$ determines the computational complexity of this method. When this ratio decreases, the amount of computation will gradually approach to the baseline, but meanwhile, the overlap region will also decrease, which reduces the quality of generated images.

To balance the amount of calculation and the quality of generated images, we let $w_{padding}$ equals to $w_{basic}$, which means $\alpha=9$. Through experiments, we found that with the decrease of $w_{basic}$ and $w_{padding}$, seamlines on the output image gradually disappeared. As shown in Fig. \ref{fig7}, the result with $w_{basic} = w_{padding} = 16$ is the best, so we use this value in subsequent experiments.

\subsection{Evaluation}

\subsubsection{Visual Evaluation}

Fig. \ref{fig8} shows the high-resolution results of our method and two aforementioned solutions, from which we observe that the results of our method are more similar to the baseline. In contrast, the results of the feathering-based method have obvious seamlines, are quite different from the baseline.

The results of these three methods at different resolutions are shown in Fig. \ref{fig9}. For a high-resolution image, the receptive field is much smaller than the image. Therefore, with the increase of resolution, more and more stylized textures are produced, which reduces the aesthetics of the generated image. However, compared with the other two solutions, our method performs better. More precisely, our method eliminates the noise textures and seamlines, which improves the quality of high-resolution stylized images.

\subsubsection{Speed Evaluation}

We tested the speed of our method on three devices: a mobile phone, a personal computer, and a GPU server. The information about these devices is as follows:
\begin{enumerate}
	\item The mobile phone is Xiaomi Mi 9, which runs on Android 10.0, powered by the Qualcomm Snapdragon 855 processor, with the Adreno 640 GPU and 8GB RAM.
	\item The personal computer runs on Windows 10, powered by the Intel Core i7-6700HQ processor, with the NVIDIA GeForce GTX 965M GPU and 4GB video RAM.
	\item  The GPU server runs on CentOS 7.0, powered by the Intel Xeon E5-2650 v4 processor, with the NVIDIA Tesla K80 GPU and 12 GB video RAM.
\end{enumerate}

On the mobile phone, we used Xiaomi's Mobile AI Compute Engine (MACE) \cite{xiaomi2018mace} to deploy models. On the personal computer and GPU server, we used Google's TensorFlow \cite{abadi2016tensorflow} to test the speed. In all tests, Fast Style Transfer models were run in GPU mode. 

\begin{table}[t]
	\footnotesize 
	\caption{Average time comparison of baseline and baseline+block shuffle (ours) for images of different resolutions on three devices.}
	\setlength{\tabcolsep}{0.7mm}{
		\begin{tabular}{@{\extracolsep{\fill}}ccccccc} \toprule
			\multirow{2}{*}{\textbf{Resolution}} &\multicolumn{2}{c}{ \textbf{Mobile Phone(s)}}& \multicolumn{2}{c}{ \textbf{Personal Computer(s)}} & \multicolumn{2}{c}{\textbf{GPU Server(s)}}\\ 
			& Baseline & Ours &  Baseline & Ours & Baseline & Ours  \\
			\midrule
			$1000^2$ & 1.23 & 16.21 & 0.91 & 8.91 & 0.72 & 5.34  \\ 
			$2000^2$ & 7.92 & 62.82 & 2.28 & 28.58 & 1.93 & 18.14 \\
			$3000^2$ & $-$ & 140.81 & $-$ & 62.96 & 3.90 & 39.27  \\
			$4000^2$ & $-$ & 246.45 & $-$ & 109.05 & 6.60 & 68.73  \\
			$5000^2$ & $-$ & 387.91 & $-$ & 168.83 & $-$ & 109.08  \\
			$6000^2$ & $-$ & 566.96 & $-$ & 244.17 & $-$ & 157.38  \\
			$7000^2$ & $-$ & 762.94 & $-$ & 339.54 & $-$ & 215.14  \\
			$8000^2$ & $-$ & 1005.76 & $-$ & 452.30 & $-$ & 280.33  \\
			$9000^2$ & $-$ & 1267.14 & $-$ & 565.95 & $-$ & 359.37  \\
			$10000^2$ & $-$ & 1525.74 & $-$ & 695.94 & $-$ & 443.07  \\
			\bottomrule
		\end{tabular}
		
	}
	\label{tab1}
\end{table}

In this experiment, we tested images with resolution ranging from $1000\times1000$ to $10000\times10000$. Tab. \ref{tab1} shows the average time for the baseline and our method to stylize images of different resolutions on the three devices, where "$-$" means the image of this resolution cannot be processed due to OOM error.

From the results, we observed that the baseline has an advantage at speed, but it can only process images with low resolution. For example, the mobile phone and personal computer can stylize images up to $2000 \times 2000$ pixels, and the GPU server can stylize images up to $4000 \times 4000$ pixels. Compared with the baseline, our method breaks through the limitation of image resolution and can stylize high-resolution images with limited memory, but at the cost of an order of magnitude slower.

\subsubsection{Memory Evaluation}
\begin{figure}[ht]
	\centering
	\includegraphics[width=8cm]{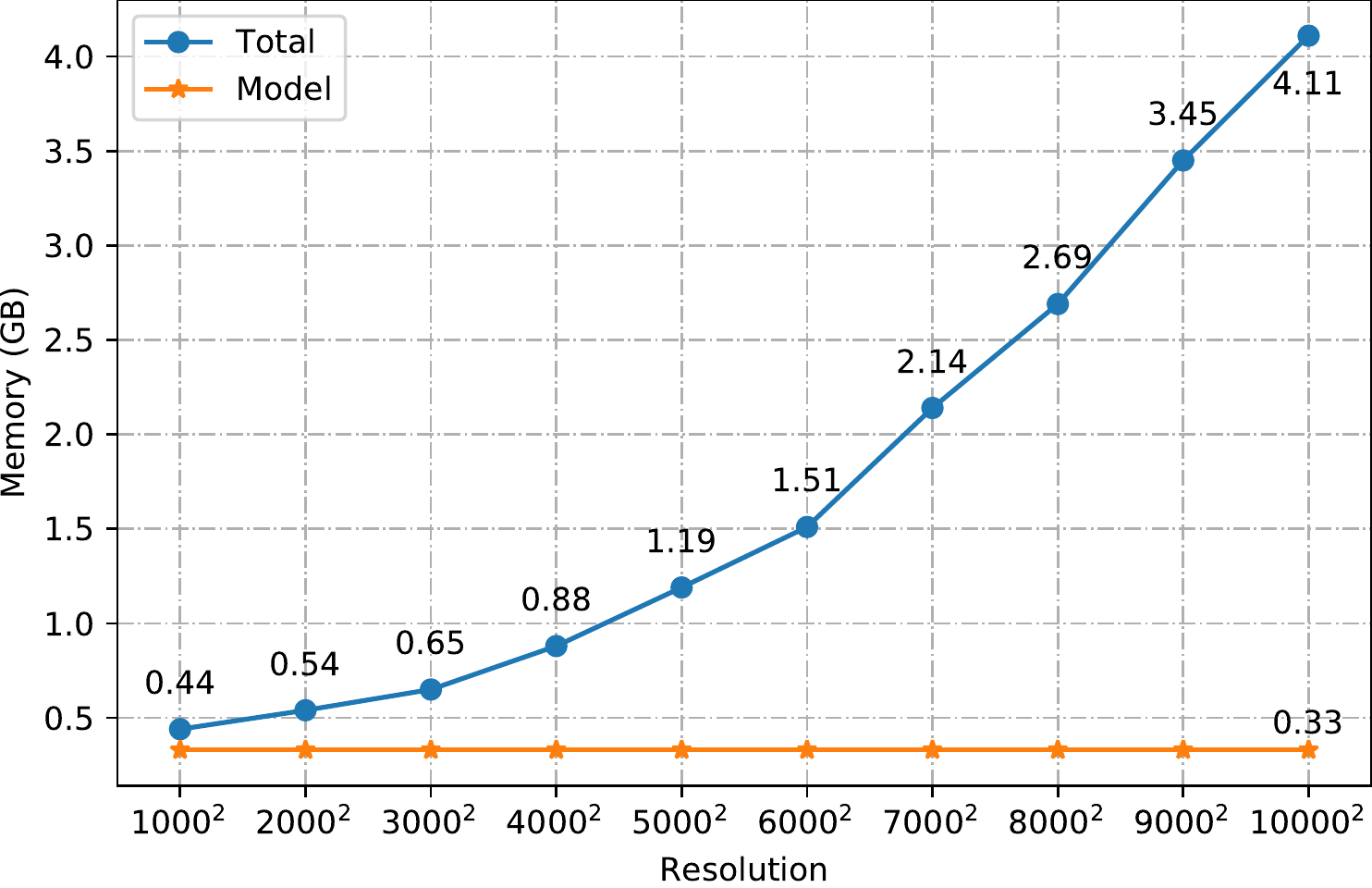}
	\caption{Memory usage (Android)}
	\label{fig10} 
\end{figure}

We tested our method with the Xiaomi Mi 9 and showed the memory usage of stylizing images with different resolutions in Fig. \ref{fig10}. From this figure, we can observe that the memory usage of the Fast Style Transfer model is a constant value of 0.33 GB. This is because the objects processed by the model are sub-images with the same resolution. 

Compared to the baseline, our method significantly reduces memory usage. More concretely, the baseline cannot stylize images above $3000\times 3000$ pixels due to OOM error, but our method even stylize a $10000\times 10000$ images will not cause OOM error. In addition, the memory usage of stylizing an image below $4000\times 4000$ pixels is less than 1 GB. It means that our method can enable most mobile devices and personal computers to support high-resolution Fast Style Transfer, which will contribute to the industrialization of Fast Style Transfer.

\begin{figure*}[!t]
	\centering
	\includegraphics[width=17.4cm]{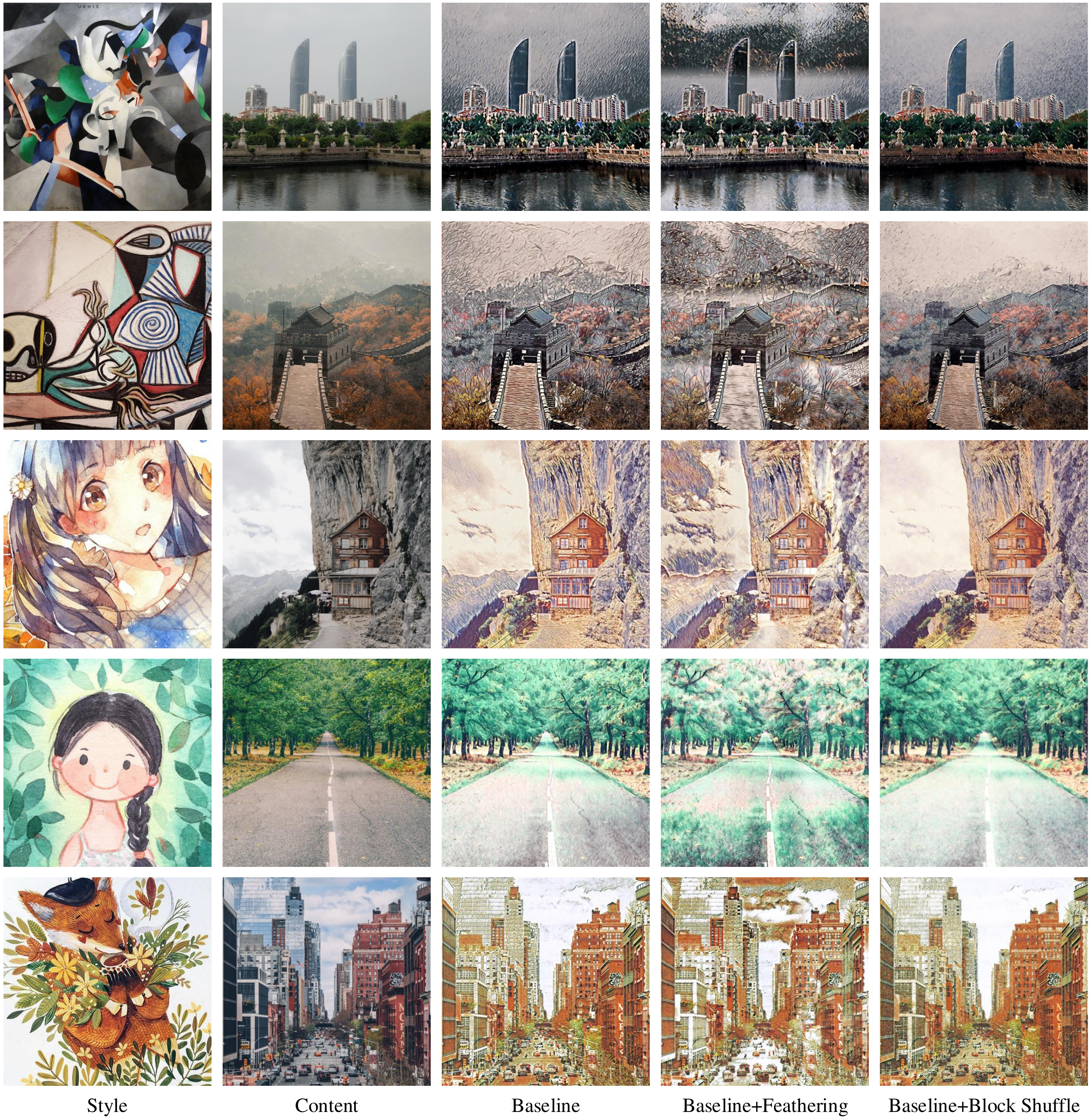}
	
	\caption{Additional comparison of baseline, baseline+feathering-based method, and baseline+block shuffle (ours). The resolution of the above images is all $3000\times 3000$.}
	\label{fig11} 
\end{figure*}

\begin{figure*}[!t]
	\centering
	\includegraphics[width=17.4cm]{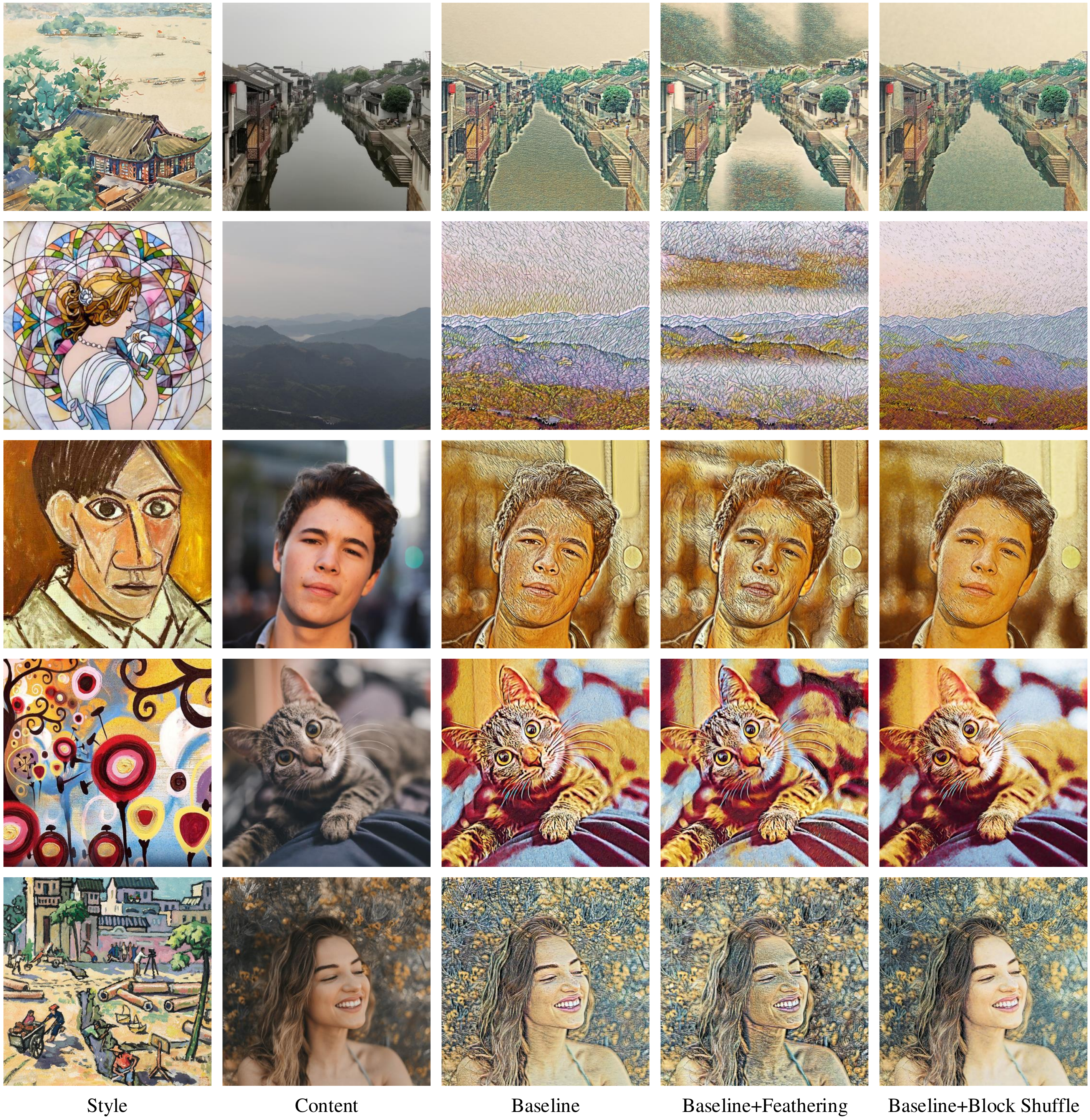}
	\caption{Additional comparison of baseline, baseline+feathering-based method, and baseline+block shuffle (ours). The resolution of the above images is all $3000\times 3000$.}
	\label{fig12} 
\end{figure*}

\section{Conclusion}
In this paper, we proposed the block shuffle method for high-resolution Fast Style Transfer with limited memory. Experiments show that the quality of high-resolution images generated by our method is superior to that of the feathering-based method. Besides, although our method is an order of magnitude slower than the baseline, it breaks through the limitation of image resolution, which enables more devices to support high-resolution Fast Style Transfer. In future work, we will further study this subject, improve the image quality and speed, and promote the industrialization of Fast Style Transfer.

\section*{Appendix A: More Comparisons}
Figure \ref{fig11} and Figure \ref{fig12} show more comparisons with baseline, baseline+feathering-based method, and baseline+blocks shuffle (ours).

\end{document}